\documentclass[10pt,twocolumn]{article} 
\usepackage{times}
\usepackage{graphicx}
\usepackage{amssymb}
\usepackage{url,hyperref}
\usepackage{color,soul}
\usepackage{adjustbox}
\makeatletter
\newcommand{\printfnsymbol}[1]{%
  \textsuperscript{\@fnsymbol{#1}}%
}
\makeatother

\date{}

\begin{document}

\title{\vspace{-1.5em}Supporting large-scale image recognition with\\ out-of-domain samples\vspace{1em}}
\author{
Christof Henkel\thanks{Both authors contributed equally.} \\
 Nvidia\\
  \texttt{chenkel@nvidia.com} \\
     \and
Philipp Singer\printfnsymbol{1} \\
 H2O.ai\\
  \texttt{philipp.singer@h2o.ai} \\

}

\maketitle

\begin{abstract}
This article presents an efficient end-to-end method to perform instance-level recognition employed to the task of labeling and ranking landmark images. In a first step, we embed images in a high dimensional feature space using convolutional neural networks trained with an additive angular margin loss  and classify images using visual similarity.
We then efficiently re-rank predictions and filter noise utilizing similarity to out-of-domain images. 
Using this approach we achieved the 1st place in the 2020 edition of the Google Landmark Recognition challenge. 

\end{abstract}

\section{Introduction}

The Google Landmark Dataset v2 (GLDv2) is a large-scale benchmark for instance-level recognition and retrieval tasks in the field of computer vision \cite{weyand2020google}. With approximately five million images spanning over 200,000 classes, several challenging data properties such as huge class imbalance, a large fraction of non-landmark test photos and intra-class heterogeneity arise. 
The third edition of the Google Landmark Recognition (GLR) challenge asked competitors to address these issues on a cleaned subset of GLDv2, which was the result of previous year's competition. The cleaned subset (GLDv2 CLEAN) consists of approximately 1.5 million images with 81,313 classes. 
While previous year's submissions were scored by uploading prediction files to the competition host, this year's follows a synchronous rerun format, where participants submit their models and code and are evaluated on an inaccessible test set. Competition entries were evaluated using Global Average Precision (GAP) \cite{perronnin2009family, weyand2020google}. This paper summarizes our winning solution to the competition. We make the code available online\footnote{\fontsize{7}{9}\selectfont{https://github.com/psinger/kaggle-landmark-recognition-2020-1st-place}}.

\section{Methodology}

\subsection{Validation strategy}

We use the test set of last year's recognition competition, which was released together with its ground truth labels \cite{cvd}, as a validation set. A crucial property of this dataset is the presence of around 98\% of out-of-domain images, which we will call non-landmarks in the following. To evaluate the GAP metric realistically we apply the same post processing step, which is applied by the host when scoring submissions: the train set is filtered to contain only classes that are also in the respective test set. We track the GAP score using softmax predictions as well as predictions derived from k-nearest-neighbors (KNN) using cosine similarity. 

\begin{figure*}[h!]
  \begin{center}
    \includegraphics[width=4.5in]{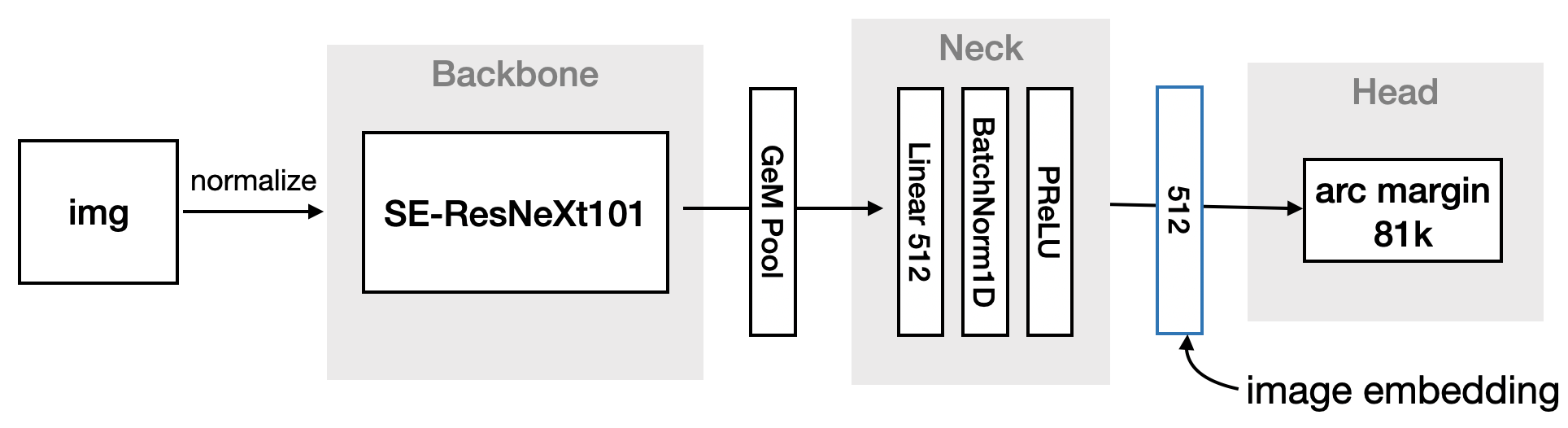}
  \end{center}
  \vspace*{-5mm}
\caption{\small Model architecture with a SE-ResNeXt101 backbone}
  \label{arch}
\end{figure*}

\subsection{Modeling}
\label{sec:modeling}

In order to efficiently distinguish a large amount of imbalanced classes, we embed images into a 512 dimensional feature space as extracted from the pooling layer of various CNN backbone models. We then match images by visual similarity by calculating their cosine similarity with known images.

To address different scales and translations, we train our models on different scales and aspect ratios with random crops. We normalize the images by the mean and standard deviation of the imagenet dataset before feeding them into a pre-trained backbone. Backbone outputs are aggregated using a Generalized-Mean (GeM) pooling layer, before feeding into a simple Linear(512) + BatchNorm + PReLU neck which outputs the 512 dimensional image embedding. We further feed the image embedding into an arc margin head \cite{deng2019arcface}  to predict one of the 81,313 landmarks. Figure \ref{arch} illustrates our setup for a SE-ResNeXt101 backbone.

Our winning submission is an ensemble  of seven models using the following ResNet \cite{he2016deep} inspired backbones SE-ResNeXt101, EfficientNet B3, ResNet152 and Res2Net101 (see Section~\ref{sec:ensembling} for details). We resize the images keeping aspect ratio as well as resize to a fixed size. In order to create diversity, the models are trained on different scales and with arc margin $m$ ranging from 0.3 to 0.4. Moreover, we either set parameter $p$ of GeM as trainable ($t$) or fixed to $3$. For inference, we use the uncropped image size which is larger than the actual training size of images as it improves the quality of extracted embeddings. Table \ref{model_ensemble_overview} gives an overview of image sizes and hyper-parameters used. 
\begin{table}[h]
\centering
\begin{adjustbox}{width=\columnwidth,center}
\begin{tabular}{|c|c|c|c|c|c|c|}
  \hline
    backbone & preprocessing & train size & test size & m & p \\
    \hline
    seresnext101 & SmallMaxSize(512) & 448x448 & 512x512 & 0.3 & t\\
    seresnext101 & SmallMaxSize(512) & 448x448 & 512x512 & 0.4 & t\\
    seresnext101 & Resize(686,686) & 568x568 & 686x686 & 0.4 & t\\
    efficientnet b3 & LongestMaxSize(512) & 448x448 & 512x512 & 0.4 & t\\
    efficientnet b3 & LongestMaxSize(664) & 600x600 & 664x664 & 0.35 & t\\
    resnet152 & Resize(544,672) & 512x512 & 544x672 & 0.4 & 3\\
    res2net101 & Resize(544,672) & 512x512 & 544x672 & 0.4 & 3 \\
    \hline
  \end{tabular}
   \end{adjustbox}
  \caption{Overview of model ensemble}
  \label{model_ensemble_overview}
 
\end{table}

\subsection{Training strategy and schedule}

We train all our models on GLDv2 CLEAN data only. Each model is trained for 10 epochs with a cosine annealing scheduler having one warm-up epoch. We use SGD optimizer with maximum learning rate of $0.05$ and weight decay of $0.0001$ across all models. We optimize using the arcface loss \cite{deng2019arcface}.

\subsection{Ranking and re-ranking out-of-domain images}
\label{sec:ranking}

As solutions to previous editions of this competition have shown \cite{chen20192nd, gu2019team, ozaki2019large}, properly ranking and re-ranking predictions is crucial to improve the GAP metric at hand that is sensitive to how landmarks and non-landmarks are ranked respectively. 
Different techniques have been proposed to tackle this task, such as (i) penalizing frequently predicted categories \cite{ozaki2019large}, 
(ii) using object detection models to find non-landmarks \cite{chen20192nd}, or use the ranked lists of test set confidences as indicators for landmarks and non-landmarks respectively. In this work, we combine ideas of previous solutions and present a holistic re-ranking routine for penalizing confidence of supposed non-landmarks.

\begin{figure}[t!]
  \begin{center}
    \includegraphics[width=0.8\columnwidth]{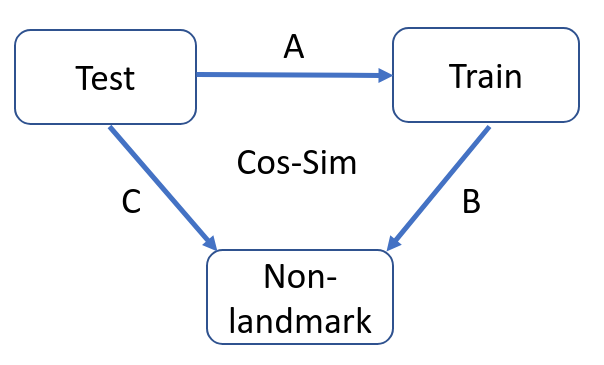}
  \end{center}
  \vspace*{-5mm}
\caption{\small Re-ranking concept}

  \label{fig:cossim}
\end{figure}

In Figure~\ref{fig:cossim}, we visualize the main concept of our re-ranking concept. Let \emph{Test} refer to the images from the test set to be recognized and ranked, \emph{Train} to the set of candidate images restricted to all possible landmarks from the test set, and \emph{Non-landmark} as an additional set of out-of-domain non-landmark images. In detail, the test set contains both unknown public and private images from the competition, train includes all images from the full GLDv2 dataset containing all images for all possible landmarks from test, and finally, non-landmark contains all images from the previous year's test set that are labeled as non-landmarks. A, B, and C refer to the all-pairs cosine similarities between respective datasets. These similarities are then utilized for finding the according landmarks identifiers and specifying the confidence in predictions.

In a general approach, one calculates the all-pairs visual similarities between test and candidate images ($A$) and then picks the landmarks with the highest similarities and uses the similarities as confidence scores. However, as mentioned, this does not penalize non-landmarks, and landmarks are not consistently ranked higher than non-landmarks which is crucial for improving the GAP metric. To that end, our approach penalizes these similarities directly by the similarity of both the candidate images and the test images against the known non-landmark images ($B$ and $C$). 
The following steps explain in detail our process for determining the appropriate landmark label and confidence for a single image $X$ in test.

\begin{enumerate}
    \item Calculate the cosine similarity between image $X_i$ and all images $Y$ from the train set ($A$).
    \item Calculate the cosine similarity between all images from train ($Y$) and all non-landmark images ($Z$). For each image in $Y$, calculate the average similarity to the top-5 most similar images from $Z$ to determine a non-landmark score for these images ($B$).
    \item Penalize the similarity $A$ by $B$, calculating $A_{i,j} = A_{i,j} - B_j$
    \item For image $X_i$, pick the top-3 most similar images from $A$, sum the similarities if images have the same landmark label in $Y$, and pick the landmark with the highest overall score and use the score as confidence.
    \item Determine the average similarity $C$ between $X_i$ and the top-10 most similar non-landmark images from $Z$ and penalize the confidence score from $A$ by subtracting $C$.
\end{enumerate}

Penalization by both $B$ and $C$ is slightly redundant; overall, penalizing the candidate images by $B$ is better, and for simplicity, one can ignore the extra step of subtracting $C$ from the final scores. One more thing to note is that the images from train, test, and non-landmark sets have slightly different distributions, and the similarity metric benefits from similarly scaled embedding vectors. To tackle this, we fitted a quantile transformers with normal distribution as output on the test set features (embedding dimensions) and applied them to the train and non-landmark datasets.  This makes the scores more stable and we assume that this also adjusts differently sized and scaled images better.

\subsection{Ensembling}
\label{sec:ensembling}

For blending our various models (see Section~\ref{sec:modeling}), we first l2-normalize each of them separately, concatenate them, and apply abovementioned quantile transformer on each embedding dimension. 
Next, we employ our ranking routine elaborated in Section~\ref{sec:ranking}. There are two ways to use the ensembled embeddings. (1) One can simply use the concatenated embedding space and run the whole procedure from start to finish on the larger embedding vectors. (2) A slightly more robust approach is to calculate $A$ and get the top-3 most similar images for each model separately, and then sum over all top-3 scores from all models. So if you have $5$ models, you would sum over $5*3=15$ and then pick the highest score.
For simplicity, the first approach works perfectly fine, reduces the complexity of the solution, and still produces accurate rankings across landmarks and non-landmarks.

\subsection{Remarks}

Interestingly, our solution does not use local descriptors, which is in contrast to previous solutions to large scale image retrieval. During the competition, we evaluated the performance of DELG \cite{cao2020unifying} and SuperPoint \cite{detone2018superpoint}, both known as state-of-the-art local descriptor models.
However, the improvement, even when using different image scales or other methods to enhance performance, was very small for both. Since the computational time for extracting and matching keypoints (especially when using DELG) was very high, we did not use local descriptors in our final submissions.
We observed that with increasing performance of global descriptors, the benefit of local descriptors diminishes and restricts to non-landmark identification. Moreover, when using the proposed method of using the global descriptor of out-of-domain images for re-ranking and non-landmark filtering, local descriptor models became obsolete.

We also would like to remark that two of our models have been initialized with backbones that were pretrained on GLD V1 \cite{gldv1}, which is a previous version of the Google Landmark Dataset. However, we later realized that this pretraining does not provide any benefit.

\section{Conclusion}

In this paper, we presented our winning solution to the Google Landmark Recognition 2020 competition. 
Our solution utilizes global features as extracted from several different backbones, fitted with an arc margin head and arcface loss.
We present a robust end-to-end ranking and re-ranking routine that efficiently positions landmark and non-landmarks in the final ordered list of predictions. It penalizes the similarity between test and candidate images by similarity against a pool of known non-landmark images. 
After ensembling several models with different backbones and training routines, we reached a final score of $0.6824$ on the public and $0.6598$ on the private leaderboard respectively.

\bibliographystyle{abbrv}
\bibliography{refs}

\begin{thebibliography}{10}

\bibitem{cao2020unifying}
B.~Cao, A.~Araujo, and J.~Sim.
\newblock Unifying deep local and global features for efficient image search.
\newblock {\em arXiv preprint arXiv:2001.05027}, 2020.

\bibitem{chen20192nd}
K.~Chen, C.~Cui, Y.~Du, X.~Meng, and H.~Ren.
\newblock 2nd place and 2nd place solution to kaggle landmark recognition
  andretrieval competition 2019.
\newblock {\em arXiv preprint arXiv:1906.03990}, 2019.

\bibitem{cvd}
{Google Landmarks Dataset v2}.
\newblock \url{https://github.com/cvdfoundation/google-landmark}.

\bibitem{deng2019arcface}
J.~Deng, J.~Guo, N.~Xue, and S.~Zafeiriou.
\newblock Arcface: Additive angular margin loss for deep face recognition.
\newblock In {\em Proceedings of the IEEE Conference on Computer Vision and
  Pattern Recognition}, pages 4690--4699, 2019.

\bibitem{detone2018superpoint}
D.~DeTone, T.~Malisiewicz, and A.~Rabinovich.
\newblock Superpoint: Self-supervised interest point detection and description.
\newblock In {\em Proceedings of the IEEE Conference on Computer Vision and
  Pattern Recognition Workshops}, pages 224--236, 2018.

\bibitem{gldv1}
{Google Landmarks Dataset v1}.
\newblock \url{https://www.kaggle.com/google/google-landmarks-dataset}.

\bibitem{gu2019team}
Y.~Gu and C.~Li.
\newblock Team jl solution to google landmark recognition 2019.
\newblock {\em arXiv preprint arXiv:1906.11874}, 2019.

\bibitem{he2016deep}
K.~He, X.~Zhang, S.~Ren, and J.~Sun.
\newblock Deep residual learning for image recognition.
\newblock In {\em Proceedings of the IEEE conference on computer vision and
  pattern recognition}, pages 770--778, 2016.

\bibitem{ozaki2019large}
K.~Ozaki and S.~Yokoo.
\newblock Large-scale landmark retrieval/recognition under a noisy and diverse
  dataset.
\newblock {\em arXiv preprint arXiv:1906.04087}, 2019.

\bibitem{perronnin2009family}
F.~Perronnin, Y.~Liu, and J.-M. Renders.
\newblock A family of contextual measures of similarity between distributions
  with application to image retrieval.
\newblock In {\em 2009 IEEE Conference on Computer Vision and Pattern
  Recognition}, pages 2358--2365. IEEE, 2009.

\bibitem{weyand2020google}
T.~Weyand, A.~Araujo, B.~Cao, and J.~Sim.
\newblock Google landmarks dataset v2-a large-scale benchmark for
  instance-level recognition and retrieval.
\newblock In {\em Proceedings of the IEEE/CVF Conference on Computer Vision and
  Pattern Recognition}, pages 2575--2584, 2020.

\end{thebibliography}
\end{document}